\documentclass[letterpaper, 10 pt, journal, twoside]{ieeetran}

\usepackage{graphicx}
\usepackage{times}
\usepackage{amsmath}
\usepackage{amssymb}
\usepackage{amsfonts}
\usepackage{array}
\usepackage{algorithm}
\usepackage{algorithmicx}
\usepackage[noend]{algpseudocode}
\usepackage{multirow,color}
\usepackage{algpseudocode}
\usepackage{varwidth}
\usepackage{subfig}
\usepackage{booktabs}
\usepackage{gensymb}
\usepackage[hyphens]{url}
\usepackage[export]{adjustbox}
\usepackage[font=footnotesize]{caption}
    
\usepackage[inline]{enumitem}
\usepackage{dblfloatfix}
\usepackage{xcolor}
\usepackage{makecell}
\usepackage{textcomp}
\usepackage{gensymb}
\usepackage{amsmath,lipsum}
\usepackage{mathtools}
\usepackage{url}
\usepackage{verbatim}
\usepackage{booktabs}
\usepackage{hhline}
\usepackage{kotex, graphicx}  
\usepackage[compress]{cite}
\usepackage[inkscapeformat=png]{svg}
\usepackage{wasysym} 

\begin{document}

\title{\fontsize{20}{24}\selectfont

Design and Validation of an Antagonistic Tendon-Driven Dexterous Robotic Hand with Bidirectional Operation}
        
\author{Chunghyeon~Lee, Hyukjun Kwon, Sungeon~Kim, Saehyun~Moon, and~Seokhwan~Jeong*


\thanks{Chunghyeon~Lee is with the Department of Mechanical Engineering,
Texas A\&M University, College Station, TX 77843, USA
(e-mail: chlee0523@tamu.edu).}%
\thanks{Hyukjun~Kwon, Sungeon~Kim, Saehyun~Moon, and Seokhwan~Jeong
are with the Department of Mechanical Engineering, Sogang University,
Seoul 04107, South Korea
(e-mail: dragonian1030@gmail.com; sungeon200102@gmail.com;
alexmoon0418@naver.com; seokhwan@sogang.ac.kr).
Corresponding author: Seokhwan~Jeong.}%
\thanks{This work has been submitted to the IEEE for possible publication.
Copyright may be transferred without notice, after which this version
may no longer be accessible.}
}


\maketitle

\begin{abstract}

Dexterous robotic hands typically reproduce human hand morphology but inherit its one-sided grasping workspace, requiring wrist or arm reorientation to grasp from the opposite side. Existing reversible hands generally rely on non-anthropomorphic, soft, or task-specific finger arrangements, whereas conventional five-digit anthropomorphic hands remain designed primarily for palmar-side grasping. This paper presents an anthropomorphic, human-scale (200~mm length), lightweight (220~g), 3D-printed, 17-DoF robotic hand built on a bidirectional antagonistic tendon-routing mechanism, in which flexion/extension (except the coupled joint) and abduction/adduction at joints are all actively driven without passive return springs. The proposed routing mechanism allows all degrees of freedom to cross their neutral configuration and form grasp closures on either the palmar or dorsal side. Experimental evaluation demonstrates an average motor-to-joint transmission error of $3.0\%$, an average  joint transmission bandwidth of $13.2$~Hz, a maximum fingertip force of $29$~N, and a positioning repeatability up to $0.15$~mm. The hand further achieves a Kapandji score of $8$, successfully performs all $33$ GRASP Taxonomy grasp types, and performs palmar- and dorsal-side grasping tasks, validating bidirectional operation in a compact, human-scale platform.

\end{abstract}


\IEEEpeerreviewmaketitle

\section{Introduction}
    
\IEEEPARstart{R}{ecent} advances in physical AI have accelerated the development of humanoid robots, and dexterous manipulation has emerged as one of the most actively pursued and closely watched challenges in this field~\cite{welte2025interactive, openai2020dexterous, billard2019trends}. In these circumstances, a robotic hand must provide a broad workspace, compact human-scale size, reliable transmission, impact compliance, and controllability within a limited volume, while remaining reasonably affordable~\cite{piazza2019century}. Because its design and transmission directly determine the available contact points and interaction dynamics with the environment, the hand is not merely an end-effector but a core embodiment of physical intelligence~\cite{gupta2021embodied}.

\begin{figure}
\centering
\includegraphics[width=0.95\linewidth]{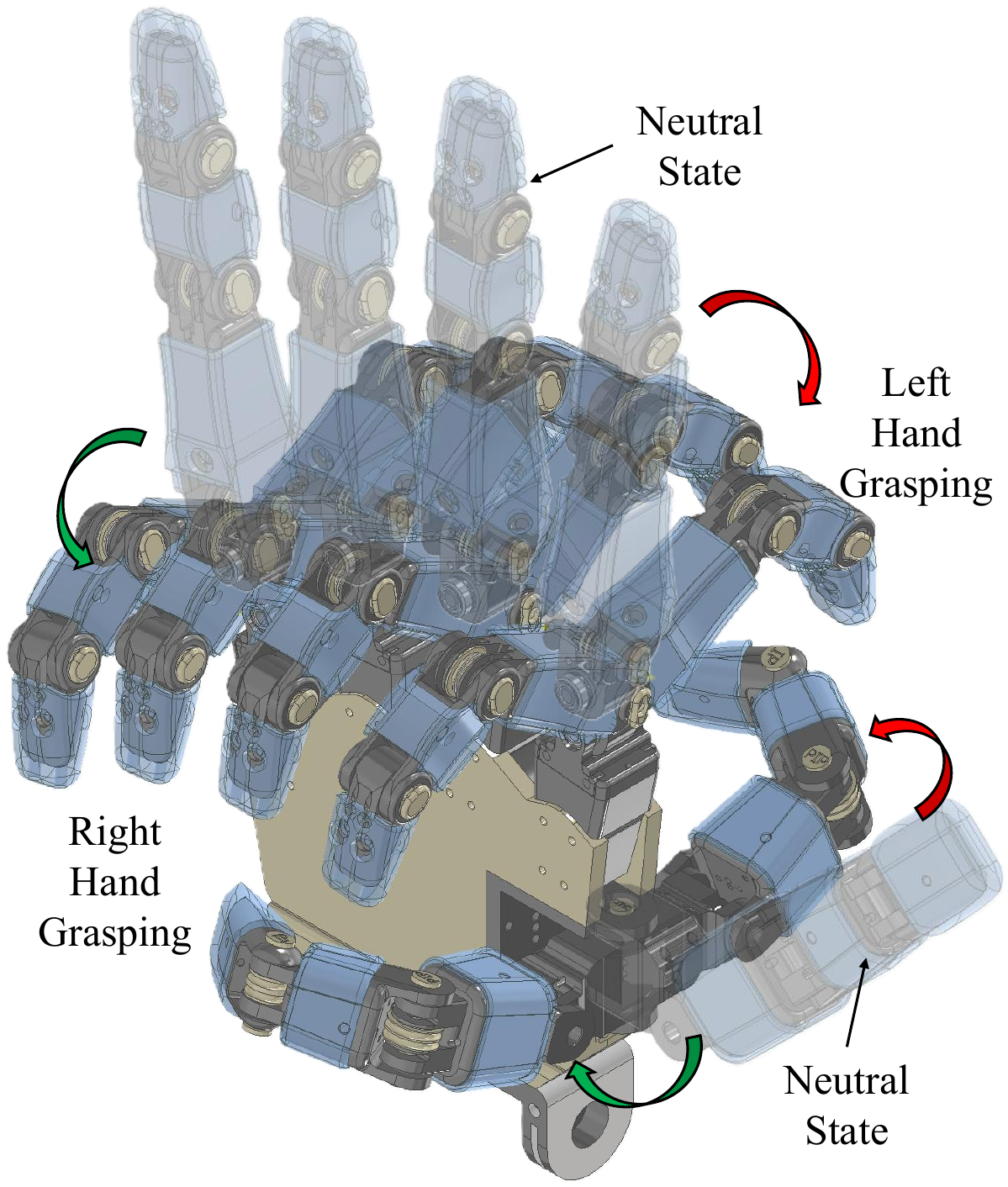}
\caption{Proposed bidirectional dexterous hand: fingers actively flex across the neutral configuration toward either the palmar (green) or dorsal (red) side, enabling bidirectional reversible two-sided grasping.}
\label{fig::fig1_HAND}
\vspace{-0.5cm}
\end{figure}

\par Most dexterous hands mimic the human hand, offering compact fingers and familiar grasp patterns. However, they inherit an anatomical asymmetry in which the fingers only bend toward the palm, making the grasping from the opposite-side inefficient due to the need to reorient the wrist or arm. Such inefficiency can be addressed if the hand supports bidirectional (or reversible) operation on either side of the palm without reorienting the wrist or arm. Examples of such bidirectional operation can be found in several systems. Gao \emph{et al.} demonstrated a reversible gripper that enables grasping on both sides of the palm, reducing wrist reorientation while expanding the workspace to handle multiple objects~\cite{gao2026detachable}. Wang \emph{et al.} developed a soft bidirectional hand in which each finger's four independently actuated chambers enable bending toward either the palmar or dorsal side, covering a wider range of grasp postures for underwater manipulation~\cite{wang2022bidirectional}. Also, Lee \emph{et al.} realized a bidirectional continuum finger via a twisted string actuator, achieving more dexterous actuation with an expanded workspace~\cite{lee2026twisted}. However, these systems rely on simplified, non-anthropomorphic fingers rather than a human-scale, multi-DoF dexterous hand.

\par Existing anthropomorphic hands can be categorized by where the actuators are placed. One common design integrates high-reduction ratio gearboxes directly within the finger joints~\cite{tesollo2026dg5fm, robotis2026hx5, allegro_hand_v4, sharpa2026wave}. This actuation offers precise position control, simple sensor integration, and, in principle, symmetric bidirectional motion. However, distributing actuators throughout the fingers and palm increases distal mass and occupies internal volume, inevitably making the hand heavier and larger. In addition, the high-reduction gears mounted directly on each joint increase reflected inertia and friction, leaving the hand less compliant to external impacts~\cite{matsuki2019bilateral}.

To move the actuators away from the fingers, tendon-driven hands relocate motors to the wrist or forearm and transmit actuation through tendons, thereby reducing distal mass and inertia. However, fully mimicking the human hand requires actively driving both flexion and extension via antagonistic tendons~\cite{zhu2022analysis}, which avoids the limitations of fixed springs and provides higher bandwidth than passive-return designs~\cite{ma2020design}. Existing tendon-driven hands illustrate this trade-off: some rely on passive spring-based return for extension~\cite{kim2019fluid}, others achieve full antagonistic actuation only at the cost of a large number of motors~\cite{grebenstein2012hand} or a high price~\cite{shadowrobot2024spec}, and others sacrifice a degree of freedom by rigidly fixing the DIP joint~\cite{christoph2025orca}. Furthermore, all of these systems lack the capability for bidirectional operation.
\par \par Therefore, this paper presents an anthropomorphic, low-cost, lightweight, tendon-driven robotic hand capable of bidirectional grasping, realized through design choices at the joint, routing, and whole-hand levels, as shown in Fig.~\ref{fig::fig1_HAND}. The main contributions of this work are as follows:
\begin{itemize}
    \item Joint design with range of motion symmetric about the neutral configuration, enabling equal actuation toward either side.
    \item Routing path design that eliminates coupling across the neutral configuration.
    \item Symmetric palmar--dorsal structure and thumb opposition, extending bidirectional operation to hand-scale grasping.
    \item Human-scale dimensions and high DoF achieved without compromising standard performance metrics despite the bidirectional design, as validated through hardware experiments.
\end{itemize}
\par The remainder of this paper is organized as follows. Section~II presents the bidirectional antagonistic tendon-routing mechanism. Section~III describes the finger, hand, and wrist/actuator design. Section~IV reports the experimental validation, and Section~V concludes the paper.


\section{Bidirectional Antagonistic Tendon-Routing Mechanism}\label{sec:Design Concept}

\subsection{Non-Thumb Finger Routing Mechanism}

\begin{figure}
\centering
\includegraphics[width=0.95\linewidth]{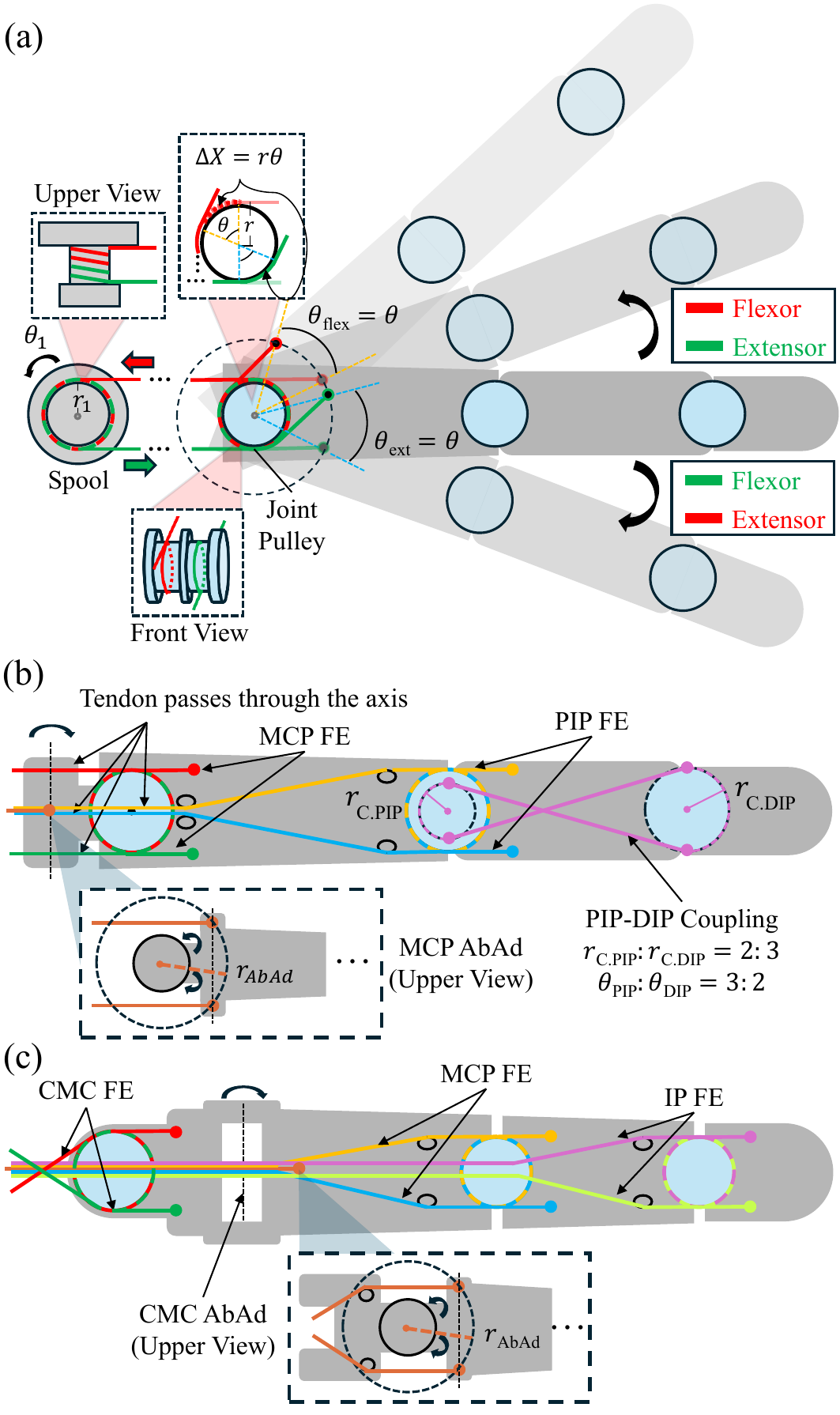}
\caption{(a) Schematic of BATM applied to the flexion/extension joint. (b) Tendon routing paths of the finger joints with human-like PIP-DIP coupling. (c) Tendon routing paths of the finger joints for active 4-DoF thumb.}
\label{fig::BATM}
\vspace{-0.3cm}
\end{figure}

The fundamental principles of Bidirectional Antagonistic Tendon-routing Mechanism (BATM) are illustrated in Fig. \ref{fig::BATM}. For assembly efficiency, identical finger modules are used for the four non-thumb fingers. Each has four DoFs, three actively driven and one passively coupled. Three motors independently actuate the two FE joints and the AbAd joint, enabling bidirectional control of all three active DoFs. However, since both flexion and extension are driven by a single motor, the variation in tendon length during extension must be equal to that during flexion. Otherwise, slack or excessive tension occurs on one side, leading to control nonlinearity or potentially causing tendon breakage.

To satisfy this constraint, the routing method shown in Fig. \ref{fig::BATM}(a) is employed. Two tendons are wound around the rotary joint in opposite directions, each capable of driving the joint in either rotational direction, so that the agonist–antagonist role of each tendon depends on the grasping direction rather than being fixed in advance. For clarity, we refer to the tendon bending the joint away from the neutral configuration as the flexor, and the one returning it to neutral as the extensor, regardless of grasping side. This configuration ensures that the following condition is satisfied:
\begin{equation}
    |\Delta X_{\text{flex}}| = |\Delta X_{\text{ext}}| = r|\theta|
    \label{eq::displacement}
\end{equation}
where $\Delta X_{\text{flex}}$ is the length variation of the flexor tendon, $\Delta X_{\text{ext}}$ is the length variation of the extensor tendon, $r$ is the radius of the rotary joint, and $\theta$ is the joint rotation angle.

Fig. \ref{fig::BATM}(b) illustrates the tendon routing path inside the finger. Joint decoupling is a primary design consideration in tendon-driven hands, as kinematic coupling between joints introduces nonlinearity during independent control of each degree of freedom. To prevent this, the metacarpophalangeal (MCP) FE tendons  are routed to pass through the AbAd axis, while the Proximal Interphalangeal (PIP) FE tendons are designed to pass through both the MCP AbAd and FE axes, ensuring axis-neutral routing at each joint. To minimize tendon friction, the routing path is composed of straight segments wherever possible. Since routing the tendons passing the AbAd and FE axes for decoupling inevitably introduces curvature, fillet geometry is applied at these regions to mitigate contact stress between the tendon and the finger structure. The PIP and MCP FE joints are routed according to the mechanism described in Fig.~\ref{fig::BATM}(a), while the MCP AbAd joint employs a straightforward routing method, approximating the tendon length variation as linear with the joint rotation angle, since the AbAd range of motion is small relative to that of FE. For the Distal Interphalangeal (DIP) joint, the PIP and DIP joints were mechanically coupled at a 3:2 ratio to generate human-like actuation.

\subsection{Thumb Routing Mechanism}

\par To enable dexterous thumb opposition, all four thumb DoFs are actively driven: FE and AbAd at the carpometacarpal (CMC) joint, FE at the metacarpophalangeal (MCP) joint, and FE at the interphalangeal (IP) joint. The corresponding tendon-routing paths are shown in Fig.~\ref{fig::BATM}(c). As with the non-thumb fingers, the CMC FE, MCP FE and IP FE joints employ the BATM, with the tendons routed past the preceding joint axes to decouple the motion of each joint.

For the CMC FE joint, the routing must satisfy two competing requirements: a compact assembly within the limited space shared with the index finger near the palm, and a sufficiently wide bidirectional range of motion to maximize the thumb-opposition workspace. To satisfy both requirements, the tendon pair was intentionally crossed before entering the palm, achieving a compact routing path while yielding a bidirectional range of motion of approximately $\pm135^{\circ}$. This wide range extends the thumb-opposition workspace across both palmar- and dorsal-side grasping configurations, increasing the number of reachable Kapandji positions and enabling successful execution of a wide range of GRASP Taxonomy grasp types. The CMC AbAd joint is configured in the same manner as the non-thumb AbAd joints and is decoupled from the CMC FE axis.


\subsection{Kinematics of the Proposed Fingers}

\begin{figure}
\centering
\includegraphics[width=1\linewidth]{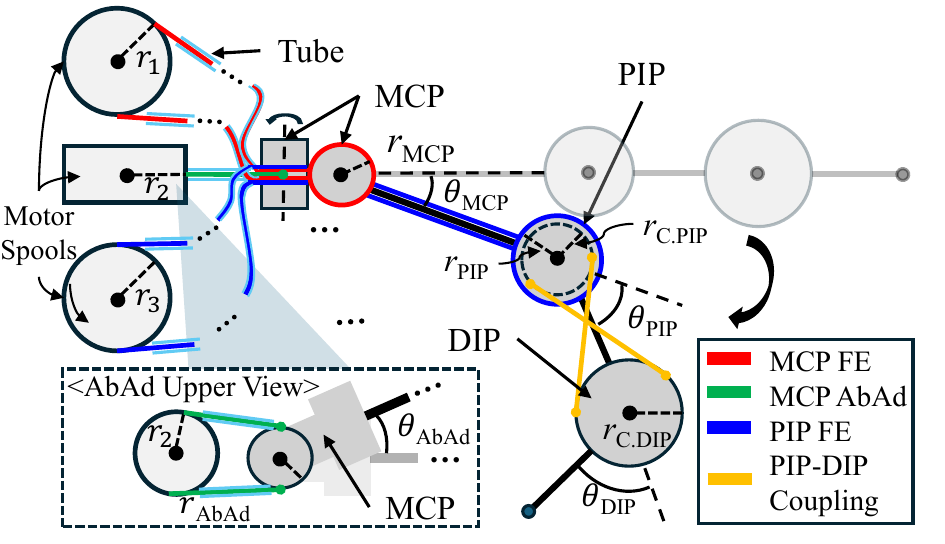}
\caption{Schematic of the finger kinematics from actuator space to joint space, illustrating the spool-to-joint transmission relationships with PIP--DIP coupling.}
\label{fig::kinematics}
\vspace{-0.5cm}
\end{figure}

As illustrated in Fig. \ref{fig::BATM}(a), the kinematics from actuator space to joint space can be modeled in a straightforward manner. Since the antagonistic tendons connected to each spool are directly coupled to the corresponding joints, the joint motion is linearly related to the spool rotation. Fig.~\ref{fig::kinematics} summarizes how each motor actuates its corresponding joint in the non-thumb fingers. As a result, the joint angle of each non-thumb finger can be expressed as a linear function of the actuator input, scaled by the ratio between the spool and joint radii, as follows.

\begin{equation}
    \boldsymbol{\theta}_{\text{finger}} = \mathbf{R} \cdot \boldsymbol{\theta}_{\text{spool}}
    \label{eq::kinematics}
\end{equation}

\begin{equation}
\resizebox{0.89\columnwidth}{!}{$
    \theta_{\text{finger}} = 
    \begin{bmatrix} 
        \theta_{\text{MCP}} \\ 
        \theta_{\text{AbAd}} \\ 
        \theta_{\text{PIP}} \\ 
        \theta_{\text{DIP}} 
    \end{bmatrix}, \quad
\mathbf{R} = 
    \begin{bmatrix} 
        \dfrac{r_1}{r_{\text{MCP}}} & 0 & 0 \\[10pt]
        0 & \dfrac{r_2}{r_{\text{AbAd}}} & 0 \\[10pt]
        0 & 0 & \dfrac{r_3}{r_{\text{PIP}}} \\[10pt]
        0 & 0 & \dfrac{2r_3}{3r_{\text{PIP}}} 
    \end{bmatrix}, \quad
    \theta_{\text{spool}} = 
    \begin{bmatrix} 
        \theta_1 \\ 
        \theta_2 \\ 
        \theta_3 
    \end{bmatrix}
$}
    \label{eq::kinematics_def}
\end{equation}

\begin{table}[htbp]
\centering
\caption{Design Parameters and Specifications of the Fingers}
\label{table::design parameters}
\resizebox{\columnwidth}{!}{
\begin{tabular}{@{}llcc@{}}
\toprule
\textbf{Definition}                      & \textbf{Symbol}        & \textbf{Value}& \textbf{Unit} \\ \midrule
MCP FE routing radius           & $r_{\text{MCP}}$                  & 4.25         & mm  \\
MCP AbAd routing radius    & $r_{\text{AbAd}}$               & 10          & mm    \\
PIP FE routing radius     & $r_{\text{PIP}}$   & 4.5          & mm    \\
CMC FE routing radius     & $r_{\text{CMC}}$   & 4.8          & mm    \\
CMC AbAd routing radius    & $r_{\text{T.AbAd}}$               & 9          & mm    \\
MCP FE routing radius for thumb           & $r_{\text{T.MCP}}$                  & 5.7         & mm  \\
IP FE routing radius     & $r_{\text{IP}}$   & 4.5          & mm    \\
PIP radius for PIP--DIP coupling                         & $r_{\text{C.PIP}}$       & 4         & mm     \\
DIP radius for PIP--DIP coupling              & $r_{\text{C.DIP}}$    & 6          & mm  \\
MCP FE joint angle            & $\theta_{\text{MCP}}$    & [-90,90]           & deg  \\
PIP joint angle            & $\theta_{\text{PIP}}$    & [-90,90]        & deg  \\
DIP joint angle             & $\theta_{\text{DIP}}$    & [-60,60]          & deg  \\
MCP AbAd joint angle                & $\theta_{\text{AbAd}}$                    & [-35,35]        & deg    \\ 
CMC FE joint angle            & $\theta_{\text{CMC}}$    & [-135,135]           & deg  \\
CMC AbAd joint angle            & $\theta_{\text{T.AbAd}}$    & [-90,90]        & deg  \\
MCP FE joint angle for thumb            & $\theta_{\text{T.MCP}}$    & [-90,90]          & deg  \\
IP FE joint angle               & $\theta_{\text{IP}}$                    & [-90,90]        & deg    \\ 
Radius of motor spools                 & $r_{i}$ \ ($i=1,\dots,4$)                  & 3        & mm    \\ \bottomrule
\end{tabular}
}
\end{table}

\begin{figure*}
\centering
\includegraphics[width=0.95\linewidth]{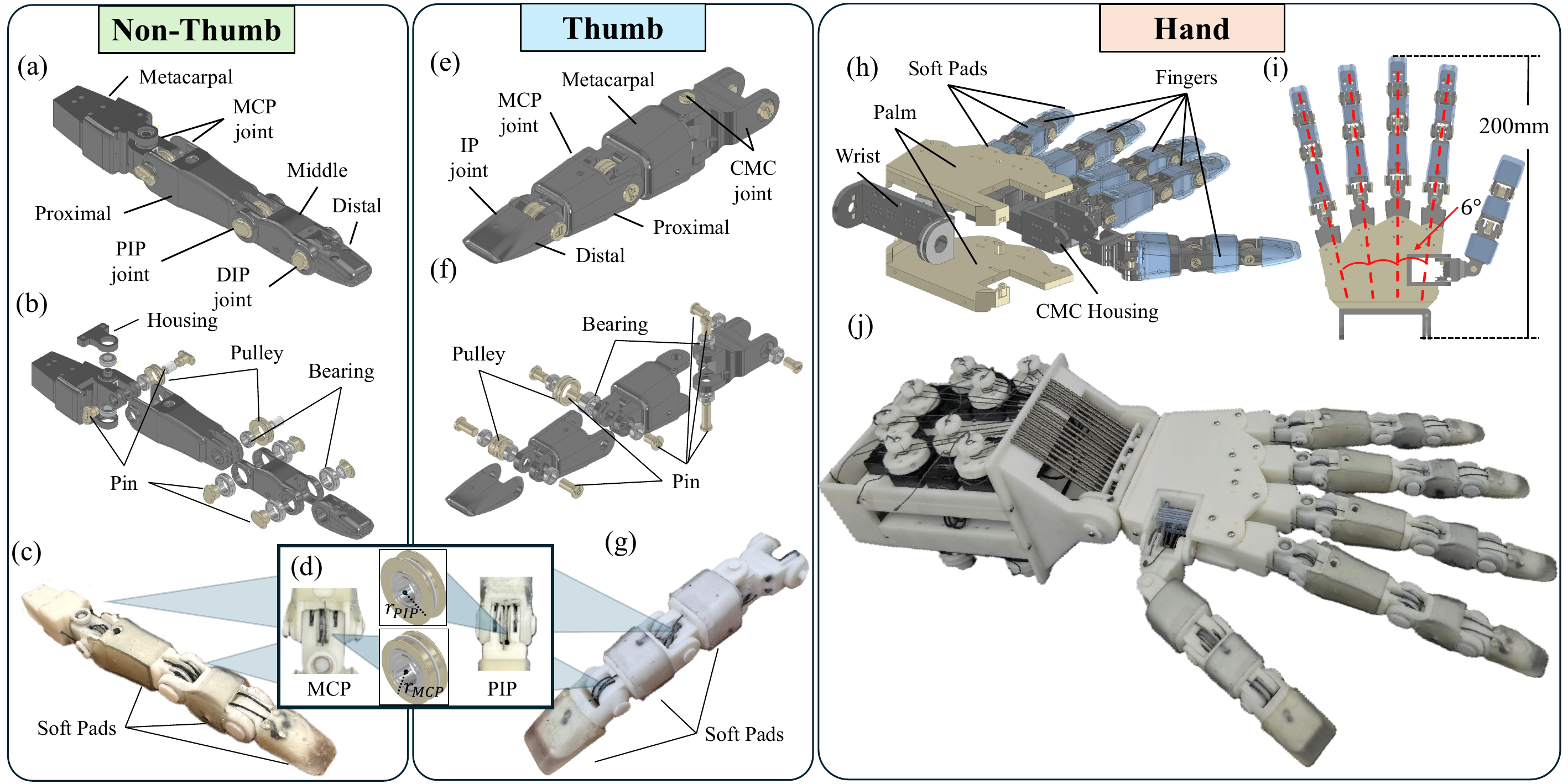}
        \caption{Design of the proposed system. Non-thumb finger: (a) assembly view, (b) exploded view, (c) fabricated prototype, and (d) pulley-bearing structure of the FE joints for friction reduction in the BATM. Thumb: (e) assembly view, (f) exploded view, and (g) fabricated prototype. Hand: (h) exploded view, (i) hand dimensions and finger splay and (j) fabricated prototype.}
\label{fig::Finger Design}
\vspace{-0.3cm}
\end{figure*}

\begin{figure}
\centering
\includegraphics[width=1\linewidth]{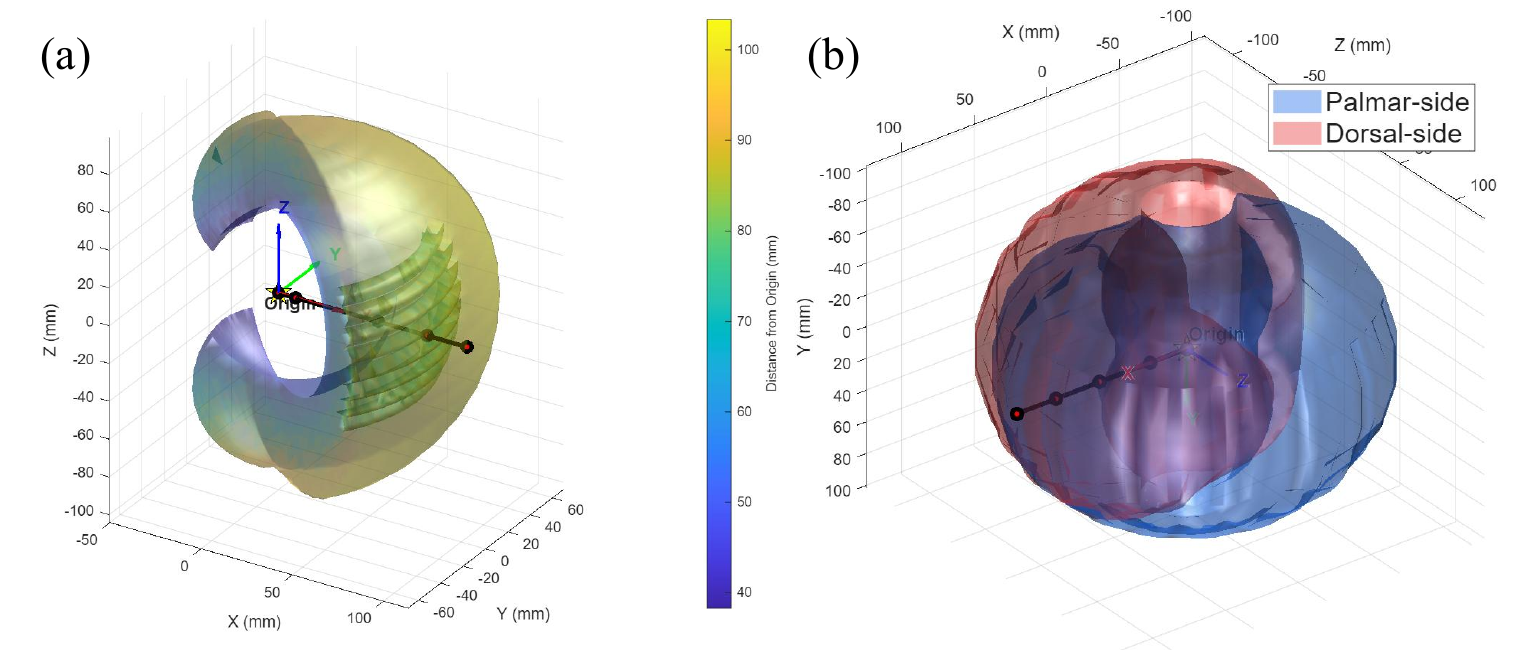}
\caption{Workspace of (a) a non-thumb finger and (b) the thumb.}
\label{fig::workspace}
\vspace{-0.35cm}
\end{figure}

\par Here, $\boldsymbol{\theta}_{\text{finger}}$ is the vector of joint rotation angles, $\boldsymbol{\theta}_{\text{spool}}$ is the vector of spool rotation angles of the motors connected to each joint, and $\mathbf{R}$ is the actuator-to-joint transmission matrix, whose entries are determined by the ratios between the motor-spool radii and the corresponding effective joint-pulley radii. The remaining geometric parameters are summarized in Table~\ref{table::design parameters}.
\par The same relationship applies to the thumb, with an additional motor extending $\boldsymbol{\theta}_{\text{spool}}$ by a fourth entry $\theta_4$ and enlarging $\mathbf{R}$ to a $4 \times 4$ diagonal transmission matrix, as follows.

\begin{equation}
\resizebox{0.89\columnwidth}{!}{$
    \theta_{\text{finger}} = 
    \begin{bmatrix} 
        \theta_{\text{CMC}} \\ 
        \theta_{\text{T.AbAd}} \\ 
        \theta_{\text{T.MCP}} \\ 
        \theta_{\text{IP}} 
    \end{bmatrix}, \quad
\mathbf{R} = 
    \begin{bmatrix} 
        \dfrac{r_1}{r_{\text{CMC}}} & 0 & 0 & 0\\[10pt]
        0 & \dfrac{r_2}{r_{\text{T.AbAd}}} & 0 & 0\\[10pt]
        0 & 0 & \dfrac{r_3}{r_{\text{T.MCP}}} & 0\\[10pt]
        0 & 0 & 0 & \dfrac{r_4}{r_{\text{IP}}}
    \end{bmatrix}, \quad
    \theta_{\text{spool}} = 
    \begin{bmatrix} 
        \theta_1 \\ 
        \theta_2 \\ 
        \theta_3 \\
        \theta_4
    \end{bmatrix}
$}
    \label{eq::kinematics_def_thumb}
\end{equation}


\section{Design of the Bidirectional Robotic Hand}

\subsection{Finger Design}
\par Based on the BATM, a finger routed only by tendons without any internal springs was designed. Fig. \ref{fig::Finger Design}(a) and (b) present the assembled and exploded views, respectively. The dimensions of the finger were modeled after the middle finger of an adult male, with phalanx lengths of 44\,mm (proximal), 27\,mm (middle) and 21\,mm (distal). Fig.~\ref{fig::Finger Design}(c) shows the fabricated prototype, 3D-printed (Bambu Lab, H2S, PLA Tough+). Also, soft pads, 3D-printed (Formlabs, Form 4, flexible 80A resin), were attached to every phalanx to increase grip friction during grasping. To minimize friction, the tendon path was kept straight wherever possible, with fillets applied at bends. However, the BATM configuration in Fig. \ref{fig::BATM}(a) has an inherent limitation: as tendon tension increases, friction at the contact interface between the joint and the tendon increases significantly. To address this, a pulley-bearing assembly was integrated at the corresponding FE joints (Fig.~\ref{fig::Finger Design}(d)), so that FE motion encounters only the minimal rolling friction of the miniature bearings. In addition, the pulley was designed with two parallel grooves, allowing the flexor and extensor tendons to route independently for bidirectional operation. Unlike conventional fingers, whose hinge structure is asymmetric to accommodate flexion toward only one side, the proposed finger adopts a hinge structure that is symmetric about the neutral configuration, allowing the joint to actuate equally in either direction.
\par For the thumb, the same materials and mechanisms as the non-thumb fingers were adopted. Fig.~\ref{fig::Finger Design}(e), (f), and (g) show the assembly view, exploded view, and fabricated prototype of the thumb, respectively.
\par Fig.~\ref{fig::workspace}(a) and (b) show the reachable workspaces of the designed non-thumb finger and thumb, respectively. Since every joint can be driven bidirectionally across its neutral configuration, each finger covers both the palmar and dorsal sides, yielding a workspace approximately twice as large as that of conventional unidirectional hands and thereby enabling more dexterous manipulation.

\subsection{Hand Design}
\par Fig.~\ref{fig::Finger Design}(h) shows the exploded view of the hand. Since the hand is designed for bidirectional, reversible operation, the fingers, palm, and dorsal structure were all made symmetric. The hand is further divided into separate parts to facilitate tendon routing within the confined internal space, as well as to simplify assembly and maintenance. Fig.~\ref{fig::Finger Design}(i) shows the overall form factor: the hand measures 200~mm, matching the size of an adult male hand, and adjacent non-thumb fingers were mounted with a $6^{\circ}$ yaw offset to obtain a fan-shaped fingertip arrangement. The distal hand weighs approximately 220~g, making it lightweight. Fig.~\ref{fig::Finger Design}(j) shows the fabricated hand prototype, whose structural components are primarily 3D-printed. In summary, the resulting system provides 17 independently actuated DoFs, including one wrist DoF, and incorporates four mechanically
coupled DIP joints.

\begin{figure}
\centering
\includegraphics[width=1\linewidth]{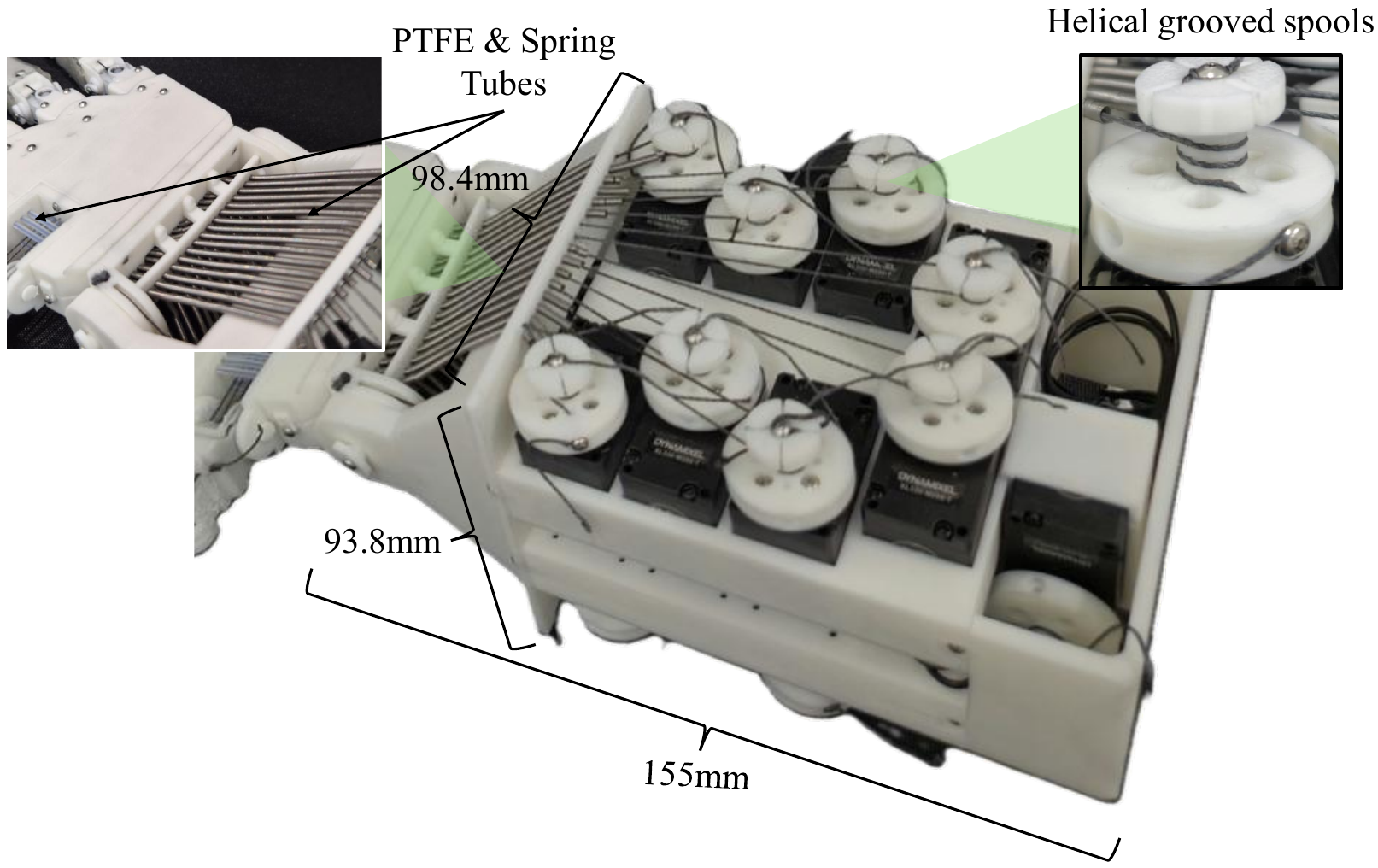}
\caption{Prototype of the wrist and actuator module, showing the spring-tube connections and helical-grooved spools.}
\label{fig::actuator_wrist}
\vspace{-0.35cm}

\end{figure}
\begin{figure}
\centering
\includegraphics[width=1\linewidth]{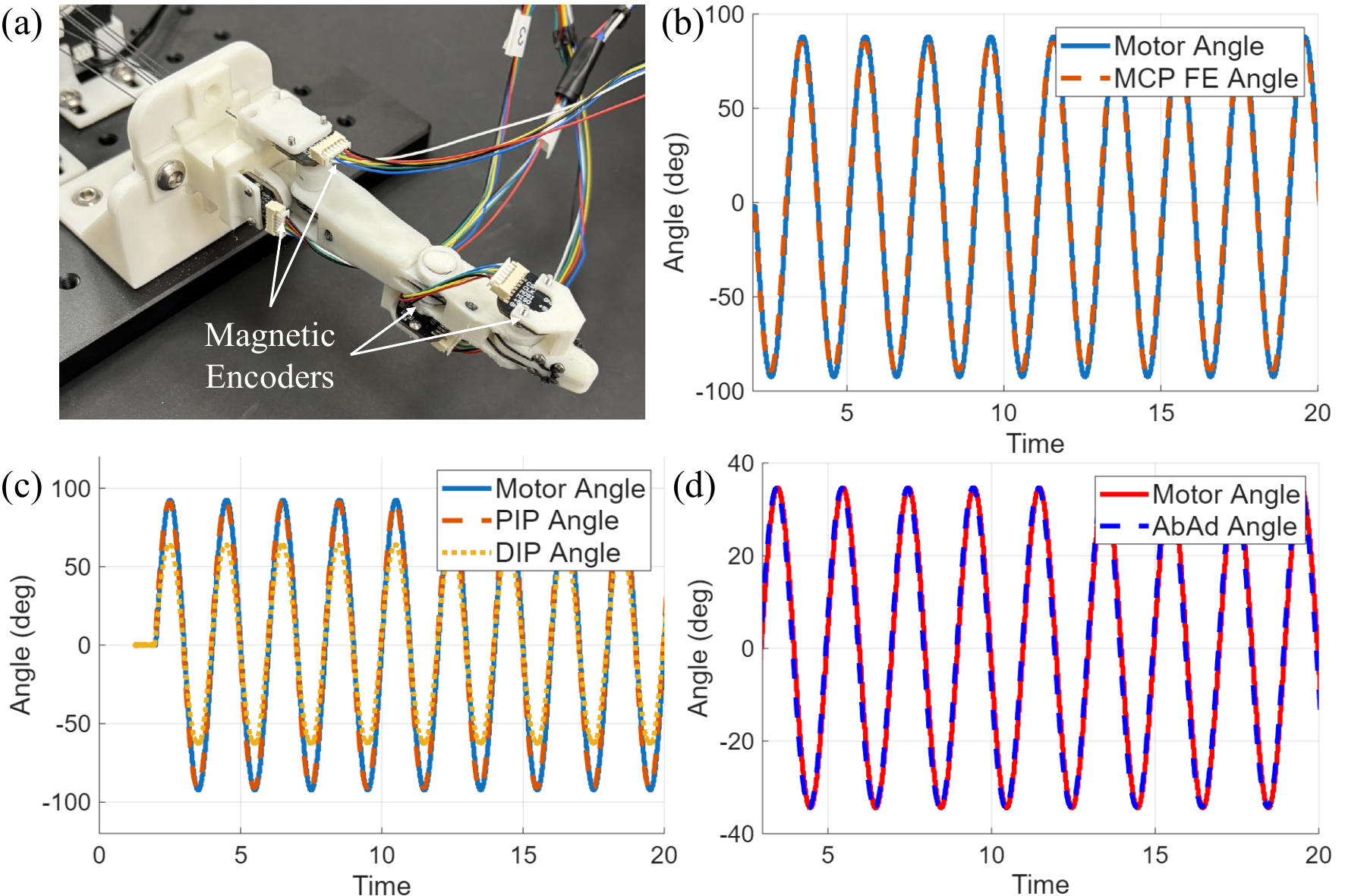}
\caption{(a) Kinematic validation setup with joint-mounted encoders. Measured and estimated joint angles for (b) MCP flexion/extension, (c) PIP/DIP flexion/extension and (d) MCP abduction/adduction.}
\label{fig::dynamic_experiment_config}
\vspace{-0.3cm}
\end{figure}

\subsection{Wrist and Actuator}
\par Fig.~\ref{fig::actuator_wrist} illustrates the wrist and actuator design. The wrist was designed with a range of motion of $-60^{\circ}$ to $60^{\circ}$. To prevent coupling between the wrist and finger tendons during this motion, all 32 finger tendons were routed past the wrist rotation axis. In addition, spring tubes were inserted along the tendon path from the hand to the actuator module, which simultaneously reduce friction and prevent unwanted variation of the tendon length. The actuator module consists of 17 motors and was designed to be compact, with overall dimensions of $98.4 \times 93.8 \times 155$~mm. In addition, each motor spool incorporates a helical groove to maintain a constant winding radius. Because the pitch was set to $1.6$~mm, the resulting deviation from the spool radius is approximately $0.36\%$ and is therefore neglected in the kinematic model.

\section{Experimental Validation}

\par The overall hardware and control setup is as follows. The tendons used are 8-strand braided Dyneema (No.\,10, TAJC Xcaribur), with a diameter of 0.6~mm and a tensile strength of 42~kgf. The motors (XL330-M288-T, Robotis) are driven via a U2D2 communication converter with a power hub board, and are controlled at a sampling frequency of 200~Hz.

\begin{figure}
\centering
\includegraphics[width=1\linewidth]{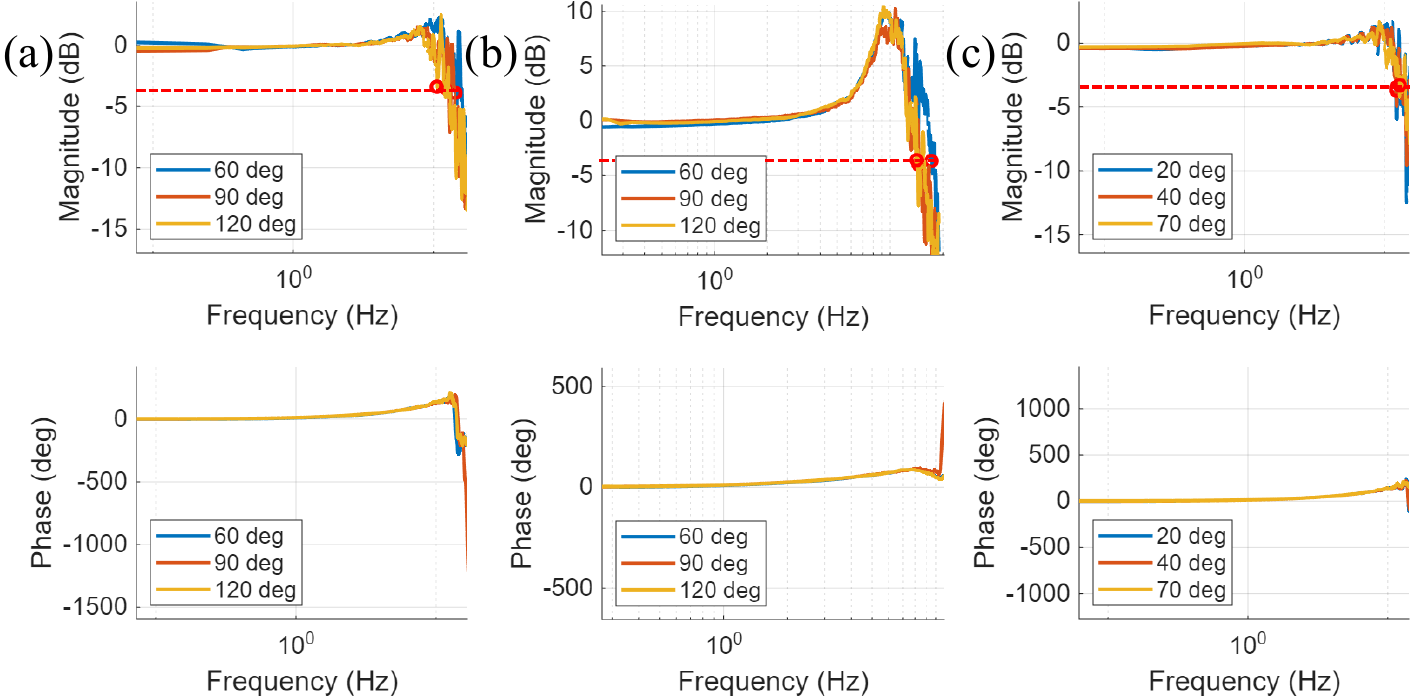}
\caption{Bode plots of joint transmission bandwidth for (a) MCP and (b) PIP flexion/extension at peak-to-peak ranges of $60^\circ$, $90^\circ$, and $120^\circ$ amplitudes, and (c) MCP abduction/adduction at peak-to-peak ranges of $20^\circ$, $40^\circ$, and $70^\circ$ amplitudes.}
\label{fig::Bandwidth}
\vspace{-0.2cm}
\end{figure}

\begin{figure}
\centering
\includegraphics[width=0.96\linewidth]{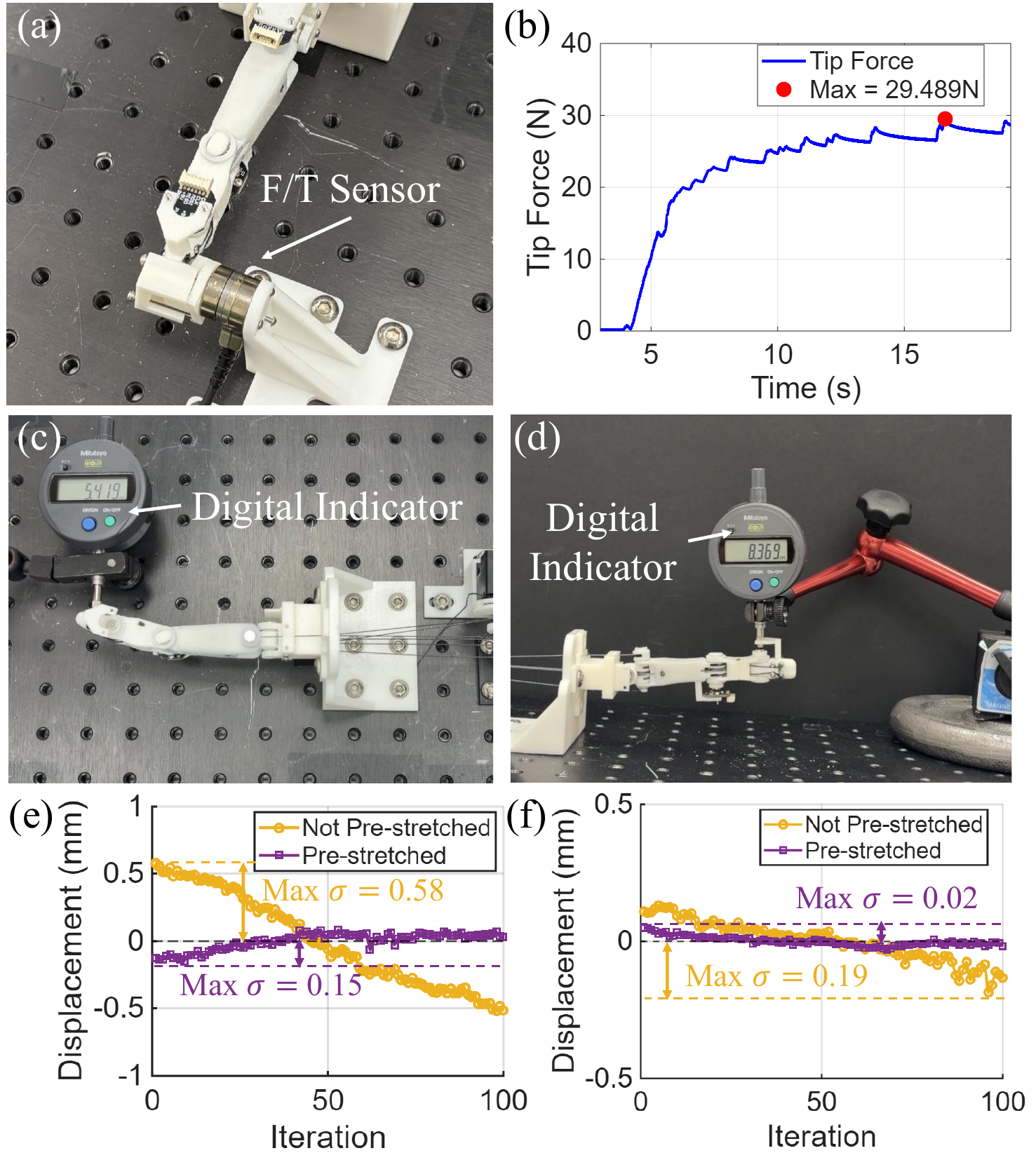}
\caption{Tip force validation: (a) experimental setup and (b) maximum tip force. Repeatability validation: experimental setup for (c) flexion/extension and (d) abduction/adduction, and repeatability results over 100 cycles with and without tendon pre-stretching for (e) flexion/extension and (f) abduction/adduction.}
\label{fig::performance}
\vspace{-0.3cm}
\end{figure}

\begin{figure*}
\centering
\includegraphics[width=0.98\linewidth]{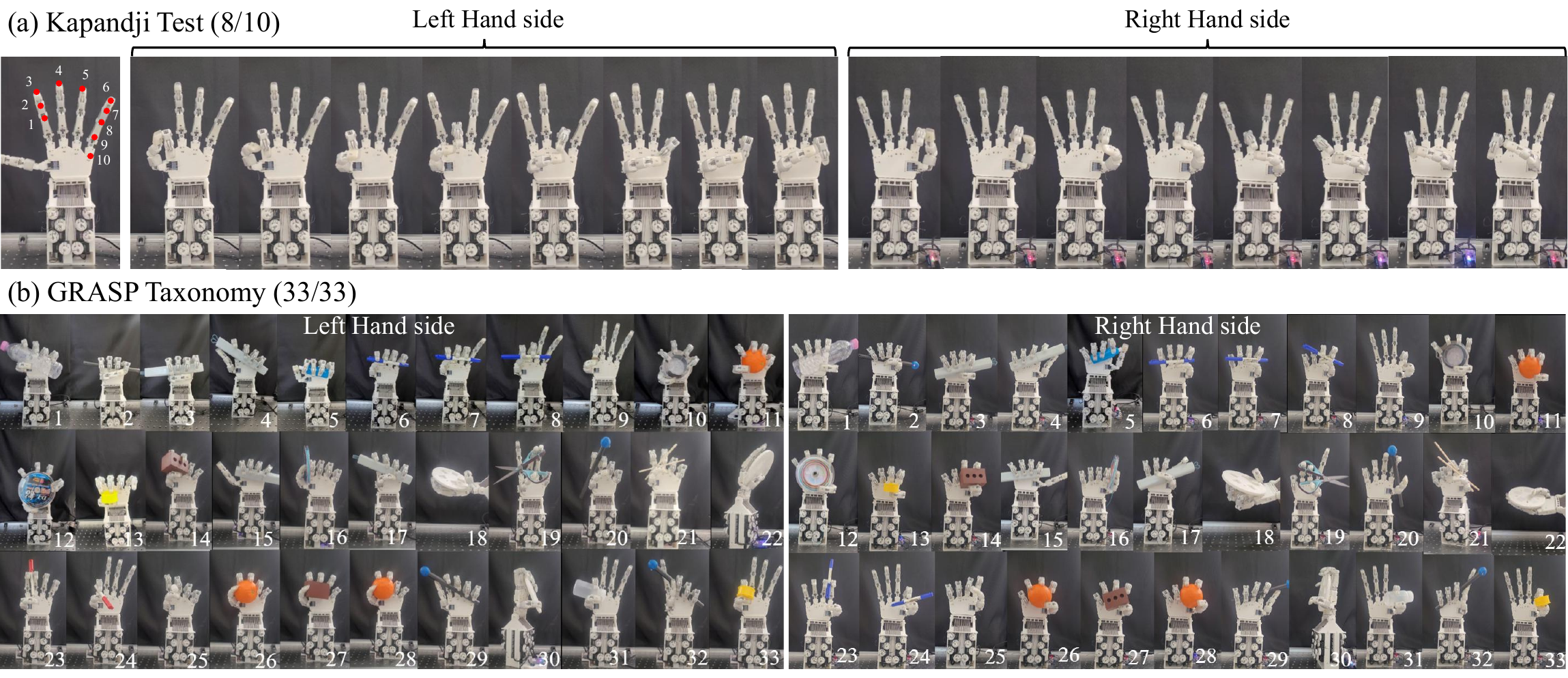}
        \caption{Dexterity evaluation using the (a) Kapandji test and (b) GRASP taxonomy, demonstrating 8/10 Kapandji positions and 33/33 successful grasp types.}
\label{fig::Hand_Test}
\vspace{-0.3cm}
\end{figure*}

\begin{figure}
\centering
\includegraphics[width=1\linewidth]{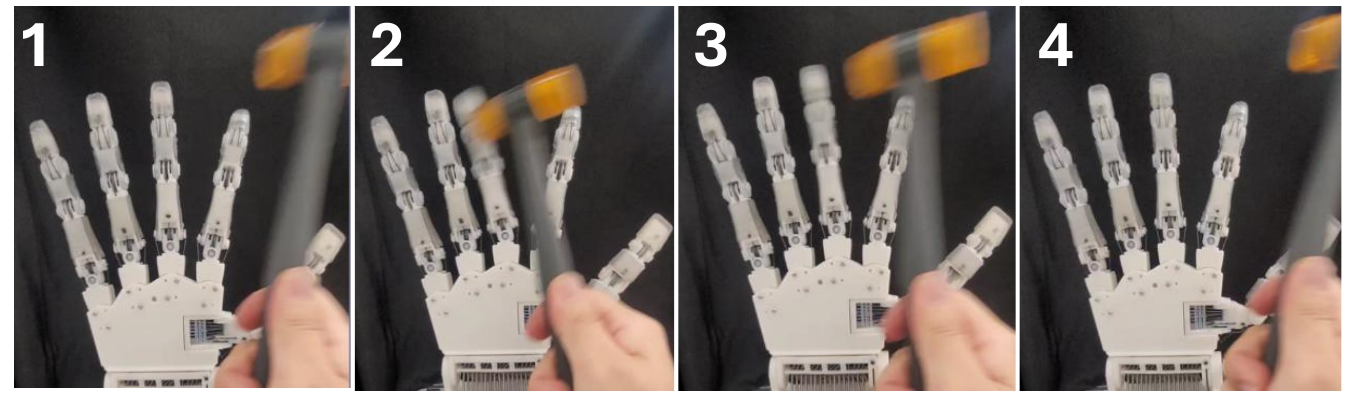}
\caption{Compliant response to external impact.}
\label{fig::impact_compliant}
\vspace{-0.4cm}
\end{figure}

\subsection{Kinematic Validation of the BATM-Actuated Finger}

\par To measure the kinematic accuracy of BATM, magnetic encoders (AS5048A, ams OSRAM) were embedded at each joint, as shown in Fig.~\ref{fig::dynamic_experiment_config}(a). Fig.~\ref{fig::dynamic_experiment_config}(b)--(d) present the validation results of the kinematic model derived in Eq.~\eqref{eq::kinematics_def}, where each joint was commanded to follow a sinusoidal trajectory over its corresponding range of motion: $-90^\circ$ to $90^\circ$ for MCP flexion/extension, $-90^\circ$ to $90^\circ$ for PIP flexion/extension, and $-35^\circ$ to $+35^\circ$ for MCP abduction/adduction. The resulting mean average errors for 10 cycles were 3.9\%, 1.6\%, and 3.4\%, respectively, demonstrating that the proposed kinematic model accurately captures the actual joint motion of BATM.

\par Linear regression on the PIP--DIP joint angle trajectories yielded a coupling ratio of 1:0.69, closely matching the target 3:2 design ratio (1:0.667) with a deviation of about 3.5\%, confirming the intended coupling behavior of the tendon-driven mechanism.

\subsection{Joint Transmission Bandwidth}
\par The joint transmission bandwidth of each joint was evaluated using the same experimental setup described above (see Fig.~\ref{fig::dynamic_experiment_config}(a)), across three different motion amplitudes, as shown in Fig.~\ref{fig::Bandwidth}(a)--(c). The input signal was defined as the measured motor angle scaled by the kinematic gain from Eq.~\eqref{eq::kinematics_def}, and the output was the joint angle measured by the encoder. The averaged bandwidths across the three amplitudes were 15.2\,Hz, 12.3\,Hz, and 12.2\,Hz for MCP FE, PIP FE, and MCP AbAd, respectively, all exceeding 12\,Hz across the entire finger. Notably, these bandwidths surpass the maximum voluntary tapping frequency of the human index finger, estimated to be approximately 6\,Hz~\cite{shourijeh2017estimation}, indicating that the proposed finger can track joint-level commands substantially faster than the natural dynamic limit of human finger movement.

\begin{figure*}
\centering
\includegraphics[width=0.95\linewidth]{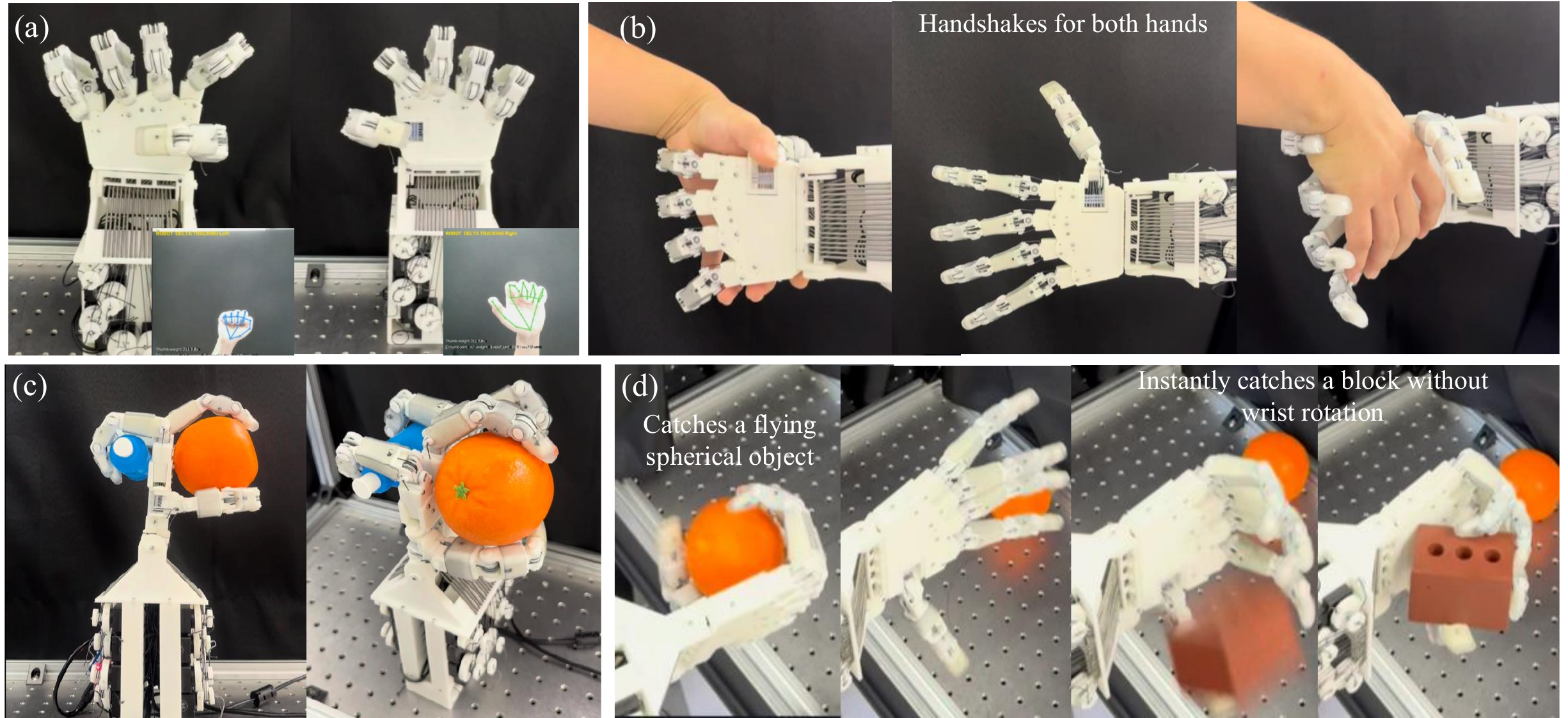}
        \caption{Applications of bidirectional operation. (a) Vision-based ambidextrous teleoperation. (b) Consecutive left- and right-handed handshakes with a single hand module. (c) Simultaneous grasping of two objects on opposite sides, exploiting the expanded workspace. (d) Dynamic grasping via bidirectional operation without wrist rotation.}
\label{fig::Bidirectional_Test}
\vspace{-0.3cm}
\end{figure*}

\subsection{Hardware Performance Validation}

\par To confirm that the bidirectional design does not come at the cost of standard hand performance, this section evaluates fingertip force, joint repeatability, impact robustness, and durability.
\par Fig.~\ref{fig::performance}(a) shows the experimental setup for measuring fingertip force, in which a 6-axis F/T sensor (Nano17, ATI Industrial Automation) was mounted at the fingertip. The motor torque was increased up to its rated value, yielding a maximum fingertip force of approximately 29~N, as shown in Fig.~\ref{fig::performance}(b). This value substantially exceeds the average maximum fingertip force of less than 15~N reported for activities of daily living~\cite{riddle2018wearable}, indicating sufficient force margin for practical manipulation tasks.

\par Fig.~\ref{fig::performance}(c) and (d) show the setups used to measure the repeatability of the FE and AbAd joints, respectively, using a digital indicator (ID-S112X, Mitutoyo, 0.001~mm resolution). The corresponding results are shown in Fig.~\ref{fig::performance}(e) and (f). Repeatability was evaluated over 100 cycles, both with and without tendon pre-stretching. For the FE joint, the maximum deviation decreased from 0.58~mm without pre-stretching to 0.15~mm with pre-stretching. For the AbAd joint, the deviation decreased from 0.19~mm without pre-stretching to 0.02~mm with pre-stretching. Based on the pre-stretched repeatability, the proposed hand achieves higher precision than existing systems, which report repeatability in the range of 0.2--1~mm~\cite{kim2019fluid, sharpa2026wave, linkerbot2026l30, agibot2026omnihand}.
\par Fig.~\ref{fig::impact_compliant} shows the finger's response to a strong impact delivered with a hammer. Owing to its tendon-driven transmission, the finger deflects compliantly with the impact and returns to its original configuration without any observed damage, demonstrating robustness to external impact. 
\par Furthermore, the finger was cyclically actuated 5,400 times over three hours without any observed failure.

\subsection{Bidirectional Dexterity Evaluation}
\par To validate the dexterity of the hand, a Kapandji test was conducted to evaluate the opposition capability of the thumb, performed independently on both the palmar and dorsal sides. Fig.~\ref{fig::Hand_Test}(a) shows the test results. Among the ten thumb opposition positions, $8$ were successfully reached, with the identical score achieved on both sides. This is within the range of 4--8 reported for existing hands~\cite{medina2024handbot,konda2023anthropomorphic,yang2025skb,zhang2026dexterous,zhou2018bcl13}. The two unreached positions are attributed to the current inter-finger spacing and thumb length; reducing the former and extending the latter would likely allow the thumb to reach the full score of 10. Nevertheless, the tip-to-tip opposition motions required for the remaining positions were performed successfully.
\par To assess the versatility of the hand in reproducing diverse grasp types, the GRASP Taxonomy was used as a validation benchmark, likewise evaluated independently on both the palmar and dorsal sides. Fig.~\ref{fig::Hand_Test}(b) shows the results: all $33$ grasp types were successfully and stably performed, with identical results achieved on both the palmar and dorsal sides.
\par The identical scores on both sides across the Kapandji test and the GRASP Taxonomy confirm that the dorsal-side workspace supports dexterous grasping to the same degree as the palmar side. Further details can be found in the supplementary video.

\subsection{Applications of Bidirectional Operation}

\par To demonstrate the practical advantages of bidirectional operation, the hand was applied to several representative applications. Fig.~\ref{fig::Bidirectional_Test}(a) shows vision-based ambidextrous teleoperation exploiting this capability: when a right hand is detected, the hand is controlled to perform right-hand like grasping, and upon switching to a left hand, the same hand module reconfigures without a hardware change to perform left-hand like grasping. Based on this, Fig.~\ref{fig::Bidirectional_Test}(b) shows the hand performing consecutive left- and right-handed handshakes using a single module. Fig.~\ref{fig::Bidirectional_Test}(c) demonstrates a task that leverages the expanded workspace: the ring and the little fingers grasp an object from one side while the remaining fingers simultaneously grasp another object from the opposite side. This would require substantial wrist or arm reorientation in a conventional palmar-only hand, whose limited workspace cannot reach both sides at once, whereas the proposed hand accomplishes it by utilizing both the palmar and the dorsal workspaces. Fig.~\ref{fig::Bidirectional_Test}(d) shows the hand dynamically catching objects thrown toward opposite sides without wrist orientation, highlighting how bidirectional operation eliminates redundant motion and enables efficient, direct catching on either side. As shown, the hand catches a flying spherical object on the palmar side and, without any wrist rotation, instantly catches a second object thrown toward the dorsal side. This demonstrates that a single hand can respond rapidly to incoming objects from either direction, without requiring two hands or wrist reorientation to switch sides. Together, these demonstrations illustrate the practical benefits that bidirectional operation brings to real-world manipulation tasks, which can translate into reductions in end-effector travel distance and task completion time compared to a unidirectional baseline. Further details can be found in the supplementary video.

\section{Conclusion and Discussion}

This paper presented a five-finger robotic hand with an overall length of 200 mm, based on the proposed BATM and providing 17 independently actuated DoFs. By actively driving both flexion/extension and abduction/adduction bidirectionally without passive return springs, the proposed mechanism allows the fingers to cross their neutral configuration and form grasp closures on either the palmar or dorsal side, extending the functional workspace beyond that of the human hand. The resulting prototype weighs approximately 220~g and was fabricated at low cost using 3D-printed components. Experimental evaluation demonstrated an average motor-to-joint transmission error of 3.0\% and a transmission bandwidth of 13.2~Hz, a maximum fingertip force of 29~N, a joint repeatability within a maximum deviation of 0.15~mm with pre-stretched tendons, and reliable operation over repeated actuation cycles, while the tendon-driven transmission allowed the hand to respond compliantly to external impact without damage. The hand further achieved a Kapandji score of 8 out of 10 and successfully performed all of the 33 GRASP Taxonomy grasp types on both the palmar and dorsal sides, and its practical benefits were demonstrated through applications including vision-based ambidextrous teleoperation, consecutive handshaking, dual-sided object grasping, and reorientation-free grasping. Together, these results validate the proposed hand as a compact, anthropomorphic platform capable of both human-like dexterity and bidirectional grasping.

Several aspects of the current design leave room for further improvement. First, the transmission bandwidth is currently constrained by the motor gear ratio and the finger weight, both of which could be further optimized. Relatedly, the current motors employ a relatively high gear ratio, resulting in a fingertip force substantially greater than necessary for typical grasping tasks; replacing them with lower-gear-ratio, quasi-direct-drive motors would improve backdrivability and enable more precise current-based force control. Second, we plan to integrate spring-based tension sensors along the tendon path to enable force and slack sensing. In addition, a passive slack-compensation mechanism would help maintain consistent tendon tension without active intervention. Finally, future work will quantitatively evaluate the advantages of bidirectional operation, such as reduced travel distance or task completion time relative to a unidirectional baseline.

\bibliographystyle{IEEEtran}
\bibliography{References}

\end{document}